# AeroBridge: Autonomous Drone Handoff System for Emergency Battery Service

Avishkar Seth[1], Alice James[1], Endrowednes Kuantama[2], Richard Han[2], Subhas Mukhopadhyay[1]

School of Engineering[1], School of Computing[2]

Macquarie University, Australia

## ABSTRACT

This paper proposes an Emergency Battery Service (EBS) for drones in which an EBS drone flies to a drone in the field with a depleted battery and transfers a fresh battery to the exhausted drone. The authors present a unique battery transfer mechanism and drone localization that uses the Cross Marker Position (CMP) method. The main challenges include a stable and balanced transfer that precisely localizes the receiver drone. The proposed EBS drone mitigates the effects of downwash due to the vertical proximity between the drones by implementing diagonal alignment with the receiver, reducing the distance to 0.5 m between the two drones. CFD analysis shows that diagonal instead of perpendicular alignment minimizes turbulence, and the authors verify the actual system for change in output airflow and thrust measurements. The CMP marker-based localization method enables position lock for the EBS drone with up to 0.9 cm accuracy. The performance of the transfer mechanism is validated experimentally by successful mid-air transfer in 5 seconds, where the EBS drone is within 0.5 m vertical distance from the receiver drone, wherein 4m/s turbulence does not affect the transfer process.

## 1 INTRODUCTION

While drone and UAV applications have recently achieved widespread popularity, one of the fundamental limitations encountered in such applications is the limited battery life of drones. Typical commercially available quadcopters and hexacopters often have a practical flight lifetime of about half an hour, after which it may take two to three hours to fully recharge their batteries. During this recharging time, the drone is unable to fulfill the mission of its application, such as monitoring, surveying, or transport. Even with advances in longevity of batteries [6], eventually the finite capacity of a drone's battery will force it to cease its duties and seek recharging.

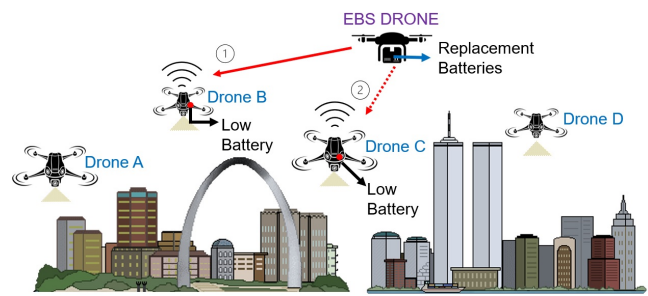

Figure 1: The AeroBridge system provides Emergency Battery Services (EBS) to in situ field drones with depleted batteries by autonomously transferring fresh batteries in mid-air.

A variety of solutions have been proposed to deal with the limited battery life of drones. Besides the current standard approach of human-assisted swapping of batteries once the drone returns to home base for recharging, another approach is to offer automatic recharging infrastructure wherein drones depart autonomously towards and land at a ground recharging station where the recharging is performed automatically [1, 5, 7, 8, 10, 25, 27, 43]. While this approach eases the task of recharging, it suffers from discontinuity in performing the drone application's mission, since the drone must break off its duties in the field in order to find a ground station to recharge. A second approach to recharging is to have the drone recharge in the air by beaming power towards the drone while in flight [9, 22]. This approach is still in its early stages and has yet to be proven practical. Another solution would be to fly a second drone to take over the application responsibilities of the first drone [45], such as monitoring or surveying, while the first drone returns to home base for recharging. This approach requires at least double the number of drones to implement, may not scale well while also proving costly, and suffers from a target re-identification problem while swapping drones [23, 29].

Instead, we explore a paradigm wherein fresh batteries are brought to drones in the field by other drones carrying the fresh batteries, and then those batteries are transferred while the drone is in mid-air. As shown in Fig. 1, this battery handoff approach, which we term Emergency Battery Services (EBS), brings the battery to the drone, rather than forcing the drone to fly back to retrieve a new battery. The EBS approach brings numerous advantages. First, the drone deployed in the field need not break off its duties of monitoring or transport in order to recharge, or at least can minimize the amount of time spent to receive a fresh supply of energy to power its motors and onboard systems. In the ideal case, the field drone never has to break contact with the subject it is monitoring and can receive the battery as it is conducting its duties, much like a runner in a relay can receive a baton while they are striding. Further, this EBS concept is powerful in that it can scale well since a single large EBS drone carrying many fresh batteries can be used to supply many drones in the field who have depleted batteries.

In addition, we envision that the EBS model in its most general form supports battery transfer that is not just one way, but two way. That is, a battery exchange or swap can occur whereby the EBS drone rendezvous with the in situ field drone and provides it with a new fresh battery, and takes away the expired battery from the field drone. This has the advantage of freeing up the dead weight of the depleted battery from having to be carried by the field drone. Many applications such as wildlife and target tracking would benefit from having effectively continuous observation. Moreover, EBS infrastructure can extend the range of deployed drones, which will be invaluable in most drone applications.

The focus and challenge of this paper will be on the computing research to solve the difficult problem of autonomously coordinating the mid-air battery handoff between two drones, namely the AeroBridge process. It does not focus on the implementation of an end-to-end battery exchange system that includes the battery connection.

**Challenges.** The first challenge is to mitigate the downdraft/downwash caused by the drones' propellers, which can considerably affect the stability of the two drones especially during handoff in which the drones are in close proximity to each other. The second challenge is relatively positioning the two drones in a stable, accurate and cost-effective manner to permit the smooth handoff of a battery from one drone to the other. This requires precise pose estimation and location estimation with an accuracy that exceeds GPS systems outdoors, and is subject to additional perturbations such as wind and imperfect motor actuation of drone propellers. We show that these effects are substantial and must be accounted for in the handoff design. Third, the challenge is to achieve the battery transfer quickly and easily to minimize disruption to the drone's primary tasks. In terms of scalability, while our system has proven effective in controlled scenarios, addressing scalability for larger drone fleets is a priority. Currently, the EBS system can handle up to three battery transfers in a single flight, but we acknowledge the need to expand this capacity for a growing number of drones. One approach is to increase the number of EBS drones to accommodate more receivers with depleted batteries. Alternatively, using larger drones with extended flight times and carrying capacity could scale operations, but we must be cautious not to exceed the limits imposed by drone size. Larger EBS drones could introduce increased downwash and potential aerodynamic disruptions to receiver drones. Additionally, the AeroBridge system is designed as an installation extension compatible with various UAV frame types, requiring minimal modifications.

In contrast to existing literature, our method uses a unique vision based algorithm to localise, position, and transfer an item to a drone mid flight. The initial location positioning is provided by GPS while the close proximity positioning uses a marker based visual odometry technique. The close proximity causes airflow disturbances which are analysed to validate the precise distance for the item transfer. This paper's main contributions consist of:

- **Downwash Effect on Drone Positioning**: We present an analysis of the downdraft/ downwash of drone propellers including modeling showing that substantial instability is caused below the propellers, making vertical transfer mechanisms difficult. Our proposed positioning method achieves a *0.5 m* drone proximity in diagonal alignment with no aerodynamic disruption. We also provide an analysis of the interactions of drone trajectory versus downwash effects.

- **Autonomous Mid-Air Docking System Design**: We propose an innovative autonomous system using two quadrotors for mid-air docking and item transfer. This approach enhances drone technology by facilitating smooth drone cooperation and positioning. Our AeroBridge system incorporates a 45-degree transfer mechanism that leverages gravity for efficient battery transfer, sidestepping downdraft instability by diagonally placing the drones.
- **CMP Model for Visual Inertial Navigation**: We implement the complete AeroBridge system, featuring fiducial marker-based pose estimation and actuated 45-degree mechanisms for battery exchange. Our unique Cross Marker Position (CMP) model optimizes visual inertial navigation by correcting orientation angles.
- **Precise Mid-Air Item Transfer Metrics**: We use the experimental evaluated readings and conduct tests to calculate exact distances, positions, and angles between two drones during mid-air item transfer.

The AeroBridge handoff coordination problem is sufficiently challenging that we focus on demonstrating one-way transfer from the EBS drone to the receiving field drone to show feasibility of the concept. We defer two-way swap to future work. In the remainder of the paper, we present related work in Section 2, investigate downwash effects on drone positioning in Section 3, describe the system architecture in Section 4, explain our cross marker position approach to visual inertial navigation in Section 5, experimentally validate the system in Section 6, and conclude in Section 7.

## 2 RELATED WORK

Several research works have explored concepts related to our mid-air battery swap idea. First, one approach employs a small secondary drone that acts as a "flying battery" that lands onto a larger primary drone, providing an additional battery source as soon as its legs contact the large drone's surface [15]. This approach has the drawback that the primary drone must carry the weight of the secondary drone since the fresh battery of the secondary drone is not separable from its body, as in our design. Also, once the primary drone's battery is depleted, then the primary drone carries the empty battery as dead weight, as the system is not conceived to support symmetric swap. Later, the authors proposed that the depleted primary battery would eventually be ejected using a parachute mechanism [17], which adds its own complications about how to recover such an ejected battery in the field. Another project seeks to lower a battery vertically from one drone to a drone below, which has a funnel to receive the lowered battery [13]. This approach is not fully autonomous, in that the top drone is flown manually by a pilot to steer the package into the funnel in the drone below.

Also, this approach suffers from the problem that vertical lowering creates a strong downwash effect [31] that causes instability in the transfer, resulting in a load that can sway, producing a pendulum effect.

Few projects have considered how to conduct in-air item transfer between drones, and not necessarily batteries. In one approach, a rod is horizontally mounted on one drone, and a second drone using a grasper grabs the rod while both are in flight [37]. This approach is limited in that this is not necessarily generalizable to other more typical payload form factors. For example, if a payload is a rectangular item such as a typical box, then the drone with the grasper arm would be forced to come much closer to the second drone and be subject to downwash-induced instability during the transfer. Also, the drone in this work is fairly imbalanced by the grasper arm, and once it possesses the rod is so imbalanced that the grasper arm is unable to keep the rod horizontal, drooping close to vertical. A second project used a string affixed to a spinning disc and payload to transfer an item between two drones but suffered from wind inaccuracies [36].

Most of the autonomous ground based battery exchange mechanisms involve the drone autonomously landing using fiducial markers [3, 26, 42, 48] and docking on a platform which swaps the charged and discharged batteries [11]. However, these methods have several disadvantages and limit the continuous operation of drones, cause significant logistics and time delays, and require complex ground stations equipped with swapping mechanisms which might be cumbersome and counter productive to flight time extension in most cases.

GPS [13] is the most commonly used in outdoor environment for most applications for localization. GPS provides location information for drones with relatively accurate positioning data, while GPS RTK [32] offers centimeter-level accuracy through ground-based reference stations. The other types of localization methods include using sensor fused data from inertial sensors along with a computer vision model [16, 47], using fiducial markers in delivery applications [35], autonomous relative navigation or docking of aircraft [21, 34, 40, 46], underwater visual-inertial localization [12], visual-inertial localization for aerial and ground robots [33], acoustic localization for landing [44], and indoor localization using acoustics [39]. The most popular uses for the ArUco markers using visual odometry are precision landing [16, 20].

Docking approaches using aerial manipulators [31], payload strings [36], payload support bar grasping [37] and robotic arm [30] are also being used for in-air manipulation tasks with payloads. While these systems work well for certain applications, ensuring a higher accuracy in docking is crucial at close proximity's to avoid instability, pendulum effects, and vibrations caused in the system.

Applying relative measurements to calculate the precise distance and orientation for the AeroBridge can be done employing a marker placed on the receiving drone. This approach works similar to a UAV vision system used in landing platforms [2, 24]. Compared to other approaches [16, 19, 28, 41] our approach uses a novel Cross Marker Positioning (CMP) technique that implements fiducial marker detection for pose estimation. This technique allows increased relative measurements and position control by implementing multiple markers placed in a custom position captured by the EBS drone.

Key studies have addressed close-proximity battery swapping [13, 15], while others have examined the nuances of proximity flight without the specific context of battery exchanges [14, 30, 31, 38, 49]. Moreover, [13] describes drones aligned vertically and hovering at a distance of 4.5 meters, employing basic PID calibration, which escalates thrust and power requirements. Furthermore, authors in [15] utilize aerodynamic disturbance analysis concepts studied in recent literature [14, 49, 50]. Analysing and experimenting the computational fluid dynamics, it is seen that a significant feedforward thrust based on the relative location of the two quadcopters is required to compensate for the proximity flight [14–16]. The secondary drone method leads to weight inefficiencies and asymmetric swapping issues [15]. Furthermore, the vertical lowering approach struggles with manual operation and downwash-induced instability [13]. Overall, the issues of aerodynamic disturbance increase with 100% overlapping of two drones, underscoring the need for precise thrust control in close-proximity flights. In summary, existing mid-air battery swap techniques face significant challenges. Thus, current methods are hindered by practical and aerodynamic limitations. In contrast, AeroBridge's angular approach emerges as a solution, minimizing downwash effects and stabilizing battery transfer, thus offering a more viable and efficient method for mid-air battery swapping.

## 3 DOWNWASH EFFECT ON DRONE POSITIONING

During the docking process, the primary objective is to analyze and determine the optimal docking angle and position, ensuring stable hovering with minimal deviation from the intended location. This necessitates thoroughly examining the aerodynamic disturbances that arise when multirotors operate in close proximity. In this context, we make the fundamental assumption that the aerodynamic disturbances encountered by a multirotor operating within the downwash of another multirotor can be predominantly attributed to two key mechanisms: The first mechanism is the drag resulting from oncoming flow on the frame. This mechanism involves the imposition of drag forces on the vehicle's frame due to the incoming airflow. This effect is a direct consequence of the oncoming flow generated by the downwash of another multirotor. The second mechanism is alteration in propeller thrust due to oncoming flow. This mechanism revolves around changes in the thrust produced by the propellers as a consequence of the incoming airflow. This alteration in thrust directly stems from the presence of oncoming flow, which is again caused by the downwash of a multirotor. It is important to emphasize that these mechanisms depend on a comprehensive understanding of the velocity field generated beneath a multirotor. We first conducted simulations in the Solidworks 3D flow simulation tool. Fig. 5 shows the drone position model based on airflow analysis at three distances. The hovering distance, $alt$, represents the height between the propellers of the receiving and EBS drone. The distance in the X axis between the two rotors is represented by $X_d$ as shown in Fig. 2. We perform a constant velocity sweep by changing $alt$ and $X_d$ values to characterize aerodynamic disturbances. We assume the receiving UAV to maintain its position setpoint. Notably, our results here are derived from experiments involving horizontal sweep simulations.

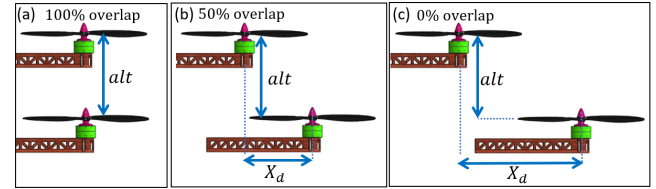

Figure 2: The drone position model based on airflow position (a) $X_d$ = 0 cm (b) $X_d$ = 16 cm (c) $X_d$ = 32 cm.

### 3.1 Airflow Estimation

Rotor-driven propellers are essential for quadcopter movement, and their analysis begins by considering the vortex effect caused by the propeller rotation. This phenomenon is exemplified in Fig. 3, where the airflow around a propeller during the vertical climb is shown for a 13-inch propeller with 5000 rpm (revolutions per minute). The downwash velocity caused is denoted by Vz [m/s]. The air pressure below the propeller is highest at the blade's center, as seen in the heat map. Based on the blade element theory [4], thrust can be calculated using mass flow rate ($\dot{m}$), thrust (T) for downstream ($T_d$) and upstream ($T_u$), and air velocity distribution (v) similarly for downstream ($v_d$) and upstream ($v_u$), as described by equations given below:

$$\begin{aligned} T &= T_d - T_u \\ &= \dot{m}(v_d - v_u) \\ &= \dot{m}[(v_d) - (v_\infty)] \end{aligned} \quad (1)$$

Now, in equation below (A) is surface area of propeller, $(v_i)$

is induced velocity and $(v_\infty)$ is free stream velocity, and air density $(\rho)$.

$$\dot{m} = \rho A[(v_\infty) + (v_i)] \quad (2)$$

The subsequent section shows the effects of the downwash from the flow rate to the drone below, highlighting unstable airflow turbulence.

## 3.2 Drone Downwash

Given the influence of downwash resulting from the presence of a nearby drone, our next objective is to employ computational fluid dynamics (CFD) to precisely determine the positioning required for the docking procedure also previously explored [14, 49, 50]. Illustratively, Fig. 3 displays simulation outcomes portraying airflow interactions between two propellers positioned at various distances, with their rotors aligned perpendicularly ($X_d$ = 0 cm). Notably, as the distance between the propellers '$alt$' increases, it becomes apparent that the effects of downwash progressively diminish. These effects are most prominent at a distance of ($alt$ = 15 cm) but notably minimal at ($alt$ = 60 cm). When juxtaposed with the reference scenario of ($X_d$ = 0 cm), diagonal displacement between the drones further reduces downwash impact.

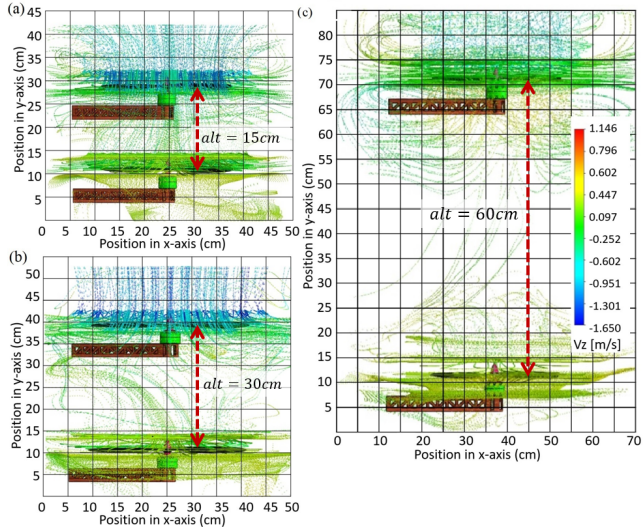

Figure 3: The airflow between two propellers with $X_d$ = 0 cm and (a) $alt$ = 15 cm (b) $alt$ = 30 cm (c) $alt$ = 60 cm.

This finding underscores the intricate relationship between relative drone positioning and the consequential mitigation of aerodynamic disturbances, offering valuable insights for optimizing mid-air battery transfer operations. Ultimately, after conducting numerous iterative simulations encompassing a range of $X_d$ and $alt$ values, the configuration yielding the optimal outcome has been identified as $X_d$ = 16 cm and $alt$ = 60 cm as seen in Fig. 4. In this specific diagonally aligned arrangement corresponding to 50 percent overlap, the disturbance caused by airflow attains its minimal magnitude.

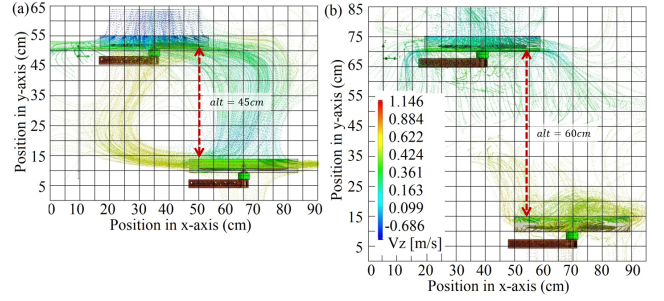

Figure 4: The airflow between two propellers with $X_d$ = 16 cm and (a) $alt$ = 45 cm (b) $alt$ = 60 cm.

## 3.3 Downwash Effects Analysis

To validate the findings of the simulations, an experimental setup was established. In the experiment shown in Fig. 5, two motorised propellers were used: one connected to a thrust logger and the other mounted on a movable platform. The objective was to assess the aerodynamic effects on the downwash-receiving (REC) propeller, as measured by the thrust logger viewed on an Arduino serial monitor. Three different overlap scenarios were examined: full overlap (100%, with $X_d$ = 0 cm), partial overlap (50%, with $X_d$ = 16 cm), and no overlap (0%, with $X_d$ = 32 cm). The thrust logger's readings were recorded under these conditions to analyze the impact of varying overlaps. Furthermore, an anemometer was used to measure the wind speeds between the two propellers as position B and behind the REC propeller as Position A. The

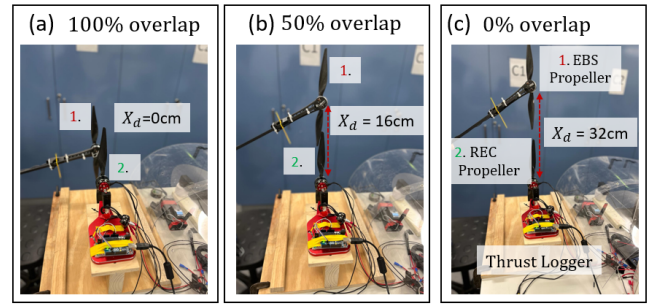

Figure 5: Thrust and Airflow Logger for varying $alt$ measurements

results obtained from the experiments, depicted in Fig. 6 (a), aimed to quantify the average airflow concerning propeller overlap and vertical distance. We placed an anemometer between the propellers of the top and bottom drones (Position

B, including grey, blue, and yellow) and also beneath the bottom propeller (Position A, including light red, green, and red) to measure how it's affected by the downwash disturbances. The experiment varied distances of *alt* between top and bottom propellers, specifically 15, 30, 45, and 60 cm, while changing $X_d$ values with 100%, 50%, and 0% propeller overlap, i.e. 100% overlap meant the two drones were directly above each other, while 50% overlap only overlapped two propellers each and 0% none. The standard average output airflow is 9.6 m/s for the drone configurations. Fig. 6 (a) validates change in output airflow due to the effects of downwash turbulence. For instance, at a 15 cm height distance, the average output airflow for 100 percent rotor overlap is 14 m/s for the bottom drone and 5.6 m/s between propellers, implying significant turbulence due to the air suction for the propellers below. Similarly, Fig. 6 (b) shows the change in thrust of the receiver drone measured due to the overlap of propellers in different positions. The propellers have an average output airflow of 9.6 m/s with 1200 gF thrust for conditions without overlap. The receiver propellers' thrust reduces due to incoming airflow disturbance, as shown by the changing *alt* values. For instance, at *alt* = 15 cm, the thrust change at 50% overlap is 200 gF, which is calculated as (1200 gF - 1000 gF). The standard average output thrust measures 1200 gF for the drone configurations; however, that is significantly reduced as observed. We see that at a vertical separation of about 60 cm, there is little change in thrust needed due to downwash, which confirms the predicted simulation results. Also, even if the two drones overlap 50%, there is little change in thrust due to downwash. For the handoff, we wish the drones to be in close proximity to increase the likelihood of a successful transfer via slides. For these reasons, we chose about 0.5 m as the vertical separation between the drones and a similar distance horizontally due to the 45° angle, which corresponds to about 50% overlap given the dimensions of the drone.

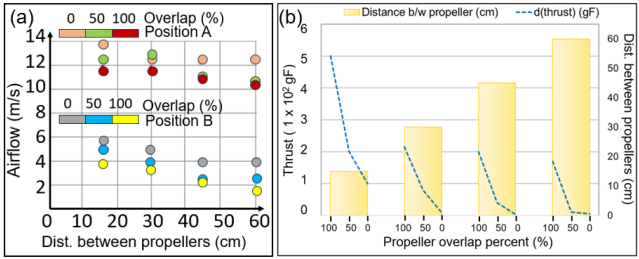

Figure 6: (a) Average o/p airflow measurements (b) Change in thrust measurement.

## 4 SYSTEM DESIGN

The AeroBridge system consists of two major dynamic components that are explained here. The first sub-system called the EBS (Emergency Battery Services) drone is used to carry and transfer the battery to the second sub-system, the 'Receiver' passive drone. To create a system architecture, two similar configuration quadrotors are used. Both drones have a carbon fiber frame of 695 *x* 695 *x* 210 mm with 13-inch propellers. The motors used in the drone are 350 KV. The system is easily re-configurable for other types of drones. The section highlights that both drones in operation have the same configurations and size. The maximum weight both drones can lift is 4.8 kg.

### 4.1 EBS Drone

We present the architecture for estimating and controlling the midair docking of two quadcopters using a dual-drone approach. The first active quadcopter, the EBS drone, is configured to carry multiple batteries. Fig. 7 shows the EBS drone designed for the proposed application. The EBS drone has an onboard computer (OBC), the NVIDIA Jetson Nano. The OBC controls the flight modes and trajectory of the drone by commanding the flight controller. Fig. 7 (a) shows that the case can carry three batteries simultaneously. The case is equipped with a servo that dispenses the battery based on the IR sensor.

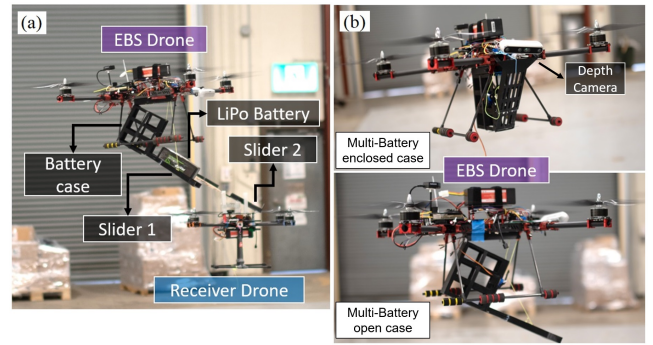

Figure 7: The battery transfer mechanism (a) Battery transfer (b)Two-stage flight of EBS drone.

The IR sensor checks which slot is available with the fully charged battery. The drone has a downward facing depth camera that enables localization, as seen in Fig. 7 (b). The 45° slide of the EBS drone is 230 mm and will attach to the receiver drone using a magnet. The receiver drone has a 45° slide that is 210 mm in length, and this allows both drones docked to be within a 0.5 m distance approximately, which is about the closest safe distance found from downwash stability analysis later. The downward-angled slide allows simplified guided gravity-assisted battery handoff and avoids complex, costly and heavy assemblages such as a robot arm [30], winch [31], or flying hot-swap battery [13] if a horizontal handoff was implemented. The entire mechanism transfers within 5 seconds. Fig. 8 shows the control mechanism of the EBS drone. It illustrates the main control components, OBC, flight controller, and microcontroller.

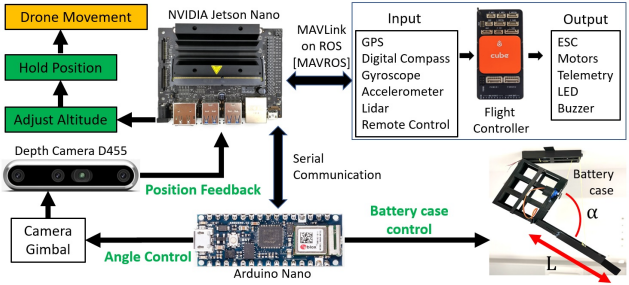

Figure 8: EBS drone mechanism.

The OBC ensures the drone's flight mode selection and positioning by communicating with the flight controller using MAVLINK. The flight controller provides flight control and GPS connectivity. The depth camera connects to the OBC, which aids the localization of the receiver drone. The microcontroller Arduino Nano 33 IoT controls the battery dispensing case; it communicates with the OBC using a UART serial communication running the 'rosserial' Arduino package. Fig. 8 also aims to highlight the system procedures for the EBS architecture. The receiver drone initiates the process by sending an emergency signal with accurate GPS coordinates. The EBS system accepts the signal request and transmits the verification signal to confirm communication. With confirmed verification, the receiver drone enables position lock to ensure stability. The EBS drone takes flight to the GPS location and begins marker screening. It detects the marker within the 3 m range of the receiver drone and commences position lock. It further adjusts the height for the optimum distance for the transfer operation. The EBS drone sends ping data to the receiver drone to allow the synchronization of the transfer slides, and both open the slides simultaneously. The slides align, then the EBS drone transfers the battery. The slides are closed on completion, and the EBS drone can return to the station, with the receiver continuing its intended mission. This structured approach ensures efficient and reliable emergency battery transfers for drones in need.

### 4.2 Receiver Drone

The AeroBridge approach assumes a scenario where an individual drone performing sensor operation, possibly in a swarm, runs out of battery. This passive quadcopter waits in HOVER mode for an EBS drone to provide an extra battery. The receiver drone has the NVIDIA Jetson Nano as the OBC to control the drone flight modes and synchronize with the EBS drone. The top base plate of the receiving drone has the Cross-Marker Position (CMP) design illustrated in Fig. 9.

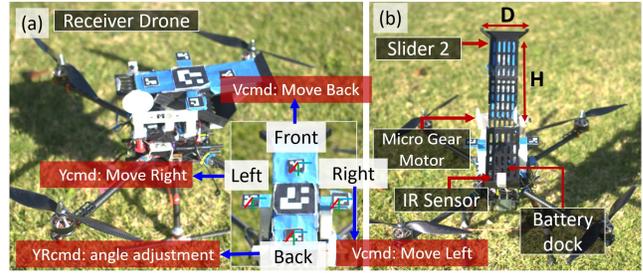

Figure 9: Receiving drone mechanism (a) CMP design (b) Drone design for the receiver.

The receiving slide is 210 mm, allowing a 500 mm distance between the two drones inclined at 45°. The receiver and EBS drone have motorized mechanisms that open and close the externally added bridges. The main reason for designing the cross-marker position is to replicate the drone's movement. The EBS drone employs an 'X' configuration structure that quickly rotates and adjusts its orientation, aligning with the front, back, left, and right fiducial markers. Fig. 9 shows the CMP design with one marker in the center with dimensions 70 x 70 mm. The central marker allows the EBS to detect the receiver drone from a 3 m distance. There are four other markers 30 x 30 mm in size; they provide a reference to the EBS drone to understand the reference position of the receiver. The micro gear motor controls the slide that extends the slide to connect to the EBS drone. The IR sensor positioned to sense a battery arriving onboard confirms receiving the battery.

## 5 CMP MODEL

This paper presents a cross-marker position localization method to determine drone positioning. This method employs the receiver drone's 6 Degree of Freedom (DoF) localization using an onboard depth camera.

### 5.1 System Block Diagram

The AeroBridge flowchart shown in Fig. 10 outlines the step-by-step process of mid-air battery transfer. It commences with the battery request from the receiver drone and proceeds to transmit relevant GPS coordinates and altitude data to the EBS drone. Upon arrival at the designated location, the EBS drone employs a marker-searching algorithm to locate the marker.

Based on marker detection success, the drone adjusts its altitude, position, and orientation for precise alignment. A handover signal triggers the opening and unfolding of the docking slides on each drone (AeroBridge), facilitating the battery transfer to the receiver drone. Following successful transfer, the docking mechanism closes on each drone, and the EBS drone returns to its home location. The flowchart concludes upon completion of the entire process. The approach for the EBS drone to approach the receiver drone in its hover position offers multiple strategies. Initially, the EBS drone leverages knowledge of the receiver drone's GPS coordinates and altitude to achieve a trajectory that minimizes total turbulence.

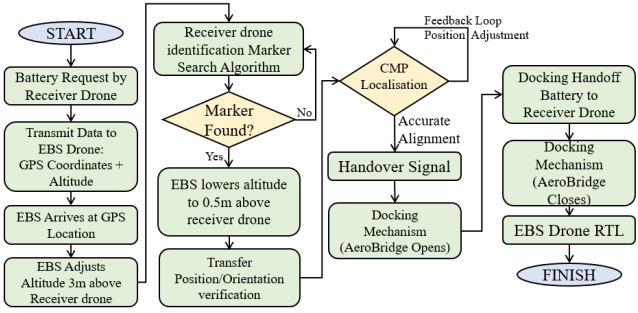

Figure 10: AeroBridge System Flow Chart

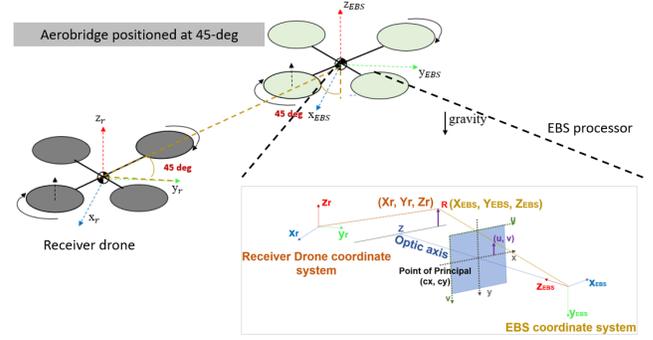

Figure 11: Receiver Drone tracker pixel coordinate system

The current operational procedure begins with placing the receiver drone in a predefined hover position with specific GPS coordinates and altitude. Subsequently, this information is relayed to the EBS drone, which conducts an initial coarse-grained search for the large marker within a 3-meter range. Upon successful detection, the EBS drone initiates a fine-grained correction, achieving a precise alignment within a 0.5-meter range using the small markers in both the horizontal and vertical dimensions for a diagonal handoff. This correction also includes orientation adjustments to ensure the EBS drone is optimally positioned for the item handoff.

## 5.2 Visual Inertial Navigation

Visual odometry estimates the relative camera motion between consecutive frames based on the tracked features. Fusing visual odometry and inertial data gives accurate and robust pose estimates. Visual-Inertial Navigation (VIN) using markers is a technique that combines information from visual data (images) and inertial measurements (accelerometer and gyroscope data) to estimate the pose (position and orientation) of a device or camera. Markers are distinctive environmental features or patterns to aid localization and tracking. The first step is detecting and tracking visual features in the camera images, such as corners, edges, or specific patterns. After successfully detecting the markers, the subsequent step is to utilize them to determine the camera's pose. Accessing the camera's calibration parameters to achieve camera pose estimation, including the camera matrix and distortion coefficient, is essential. The EBS coordinate system is developed to verify the position of the drone. The camera is positioned at 45° on the drone in flight, based on experiments discussed later, to understand this setup. The aim is to estimate the pixel coordinate for the receiver drone and depict the deviation in position using the CMP model. The authors consider three coordinate systems for this positioning system: The receiver-drone coordinate system, EBS coordinate system, and the frame coordinate system. The drone coordinate system helps determine the positions of points in the flight facility. The coordinate system selects an origin (0, 0, 0) by choosing a corner and defining the X, Y, and Z axes along the ground and vertical dimensions. With this setup, any drone location in the GPS location is in 3D space by measuring its distance from the origin along the X, Y, and Z axes. The drone coordinate system is represented in Fig. 11 using orange-colored axes. For instance, if we consider the receiver drone R in the room, its coordinates in the drone Coordinate System would be represented as $X_r, Y_r, Z_r$. To transform the coordinates of point R from the drone coordinate system to the EBS coordinate system, we apply the rotation matrix $R$ and the translation vector $t$. This process expresses the coordinates of point R in the camera's coordinate system as $X_{EBS}, Y_{EBS}, Z_{EBS}$. Mathematically, the transformation equation (3) is:

$$\begin{bmatrix} X_{EBS} \\ Y_{EBS} \\ Z_{EBS} \end{bmatrix} = (R) \begin{bmatrix} X_r \\ Y_r \\ Z_r \end{bmatrix} + t \qquad (3)$$

Once the point R is in the camera's coordinate system, it can be projected onto the image plane using equations derived from similar triangles. The extrinsic matrix combines the rotation and translation vector to transform the 3D point from the drone coordinate system to the camera coordinate system, as seen in Eq. (4). The EBS coordinate system projects the 3D point $X_{EBS}, Y_{EBS}, Z_{EBS}$ onto the image plane. This projection results in the image plane's 2D coordinates x and y.

$$E = [R|t] \quad (4)$$

The intrinsic matrix (I) Eq. (5) represents the process containing the camera's intrinsic parameters. The camera's intrinsic parameters, like the focal length (f), are used in the intrinsic matrix to project 3D points onto the image plane; this allows for a more convenient representation and calculation of the image coordinates.

$$I = \begin{bmatrix} f_x & \alpha & c_x \\ 0 & f_y & c_y \\ 0 & 0 & 1 \end{bmatrix} \quad (5)$$

The camera's optical center, represented as c_x and c_y, may not coincide with the center of the image coordinate system. This offset indicates that the camera's principal point (the point where the optical axis intersects the image plane) is not at the center of the image. A slight skew angle $\alpha$ may exist between the x and y axes of the camera sensor; this means that the axes are not perfectly perpendicular to each other, leading to a slight rotation in the image. The intrinsic matrix (I) accounts for these parameters, allowing for an accurate projection and transformation from 3D world coordinates to 2D image coordinates.

### 5.3 CMP Localization

For this application, the markers used are the '4X4 DICT' (i.e. the 4x4 dictionary range of fiducial markers with a 4x4 black-and-white squares internal grid). Each receiver drone will be equipped with a unique marker design, i.e., the Cross Marker Position (CMP). The Fig. 9 illustrates the CMP configuration and the detection of markers with position and orientation values. The CMP method references five positions in the marker; the front, back, left, right, and center. The CMP method detects the position and ID of all five markers. The EBS drone is equipped with a depth camera that will detect the position and ID of the markers. The camera has been calibrated to resolve any distortion or depth estimation. The position localization is determined using Robot Operating System (ROS). The algorithm shows the syntax for the arguments defined for the launch script, and this can be modified based on the application.

The developed package determines the marker frame and the reference position with respect to the camera alignment. The aruco_pose ROS topic will publish the pose and the orientation. The CMP method requires simultaneous detection of five markers and the default reference frame to allow the EBS drone to navigate. The previous section discusses the use of four markers 0.03 m and one 0.07 m marker, all derived from of the '4X4' DICT. The drone positions itself with 'vel_cmd_y' velocity commands for moving left and right, and uses 'vel_cmd_x' to move front and back. The 'yaw_rate_cmd' allows the EBS drone to adjust its yaw angle orientation. The authors in [18] estimate angular velocities that can be determined using Euler rates. The equation 6 shows the correlation between drone angular velocities (p, q, r) with the angular rotations values of pitch, roll, and yaw represented as $\phi$, $\theta$, and $\psi$, respectively.

---

**Algorithm 1** CMP Localization

1: **GPS Location and Altitude measurements**
2: Marker Dictionary Configuration
3: Locate Receiver Drone
4: **while** Receiver drone detected **do**
5:   arg name=marker_size default=0.07
6:   **for** Marker Position (Front Left Right Back) **do**
7:     arg_name=tf_prefix default=marker_id
8:     velocity_ned = back_position − drone_position
9:     Orientation Estimation $yaw\_rate\_cmd$
10:    **if** Position Accurate **then**
11:       EBS Position Lock
12:    **end if**
13:  **end for**
14: **end while**
15: **Output**: Open Aerobrige for transfer

---

$$\begin{bmatrix} p \\ q \\ r \end{bmatrix} = \begin{bmatrix} \phi - \psi sin\theta \\ \theta cos\phi - \psi cos\theta sin\phi \\ \psi cos\theta cos\phi - \theta sin\phi \end{bmatrix} \quad (6)$$

To estimate the quaternion for drone movements containing four real parameters, they are propagated according to the differential equation (7). The matrix form illustrates a product between the angular rates and quaternion values represented as $\dot{x}\ \dot{y}\ \dot{z},\ \dot{w}$. These values are extracted using the CMP localization on the receiver drone.

$$\begin{bmatrix} \dot{x} \\ \dot{y} \\ \dot{z} \\ \dot{w} \end{bmatrix} = \frac{1}{2} \begin{bmatrix} -y & -z & -w \\ x & -w & z \\ w & x & -y \\ z & y & x \end{bmatrix} \begin{bmatrix} \phi - \psi sin\theta \\ \theta cos\phi - \psi cos\theta sin\phi \\ \psi cos\theta cos\phi - \theta sin\phi \end{bmatrix} \quad (7)$$

To calculate the accurate positioning for the EBS drone, the authors propose equation 8. The EBS drone will estimate the appropriate orientation and position of the drone using the product of the cosine (c) and sin (s) values of GPS latitude (la) and longitude (lo) values that are represented in the matrix factoring the radius of the earth (R) and the quaternion matrix derived using CMP localization.

$$CMP_Loc = \begin{bmatrix} Rc(la)c(lo) & 0 & 0 \\ 0 & Rc(la)s(lo) & 0 \\ 0 & 0 & Rs(la) \end{bmatrix} \begin{bmatrix} \dot{x} \\ \dot{y} \\ \dot{z} \\ \dot{w} \end{bmatrix} \quad (8)$$

This system can be translated to any drone size as the algorithm can be modified to accommodate the preferred marker size.

## 6 RESULTS AND VALIDATION

### 6.1 Displacement Validation

The decision about the position of the two drones is imperative to achieve a balanced transfer. While hovering in a fixed position, the multi-rotor generates gales known as "downwash." To investigate the effect of downwash, we had one UAV fly beneath another UAV hovering in a fixed place in our experiment. These experiments validate the destabilization of the drone due to air suction in the drone below. A motion-tracking system using four external beacons was positioned to verify the drone's location. Fig. 12 shows the three stages of the drone flight wherein the beacons 10 and 11 represent the EBS and Receiver drone, respectively. The table in the figure shows the distance of the drones from each beacon, and the green-coloured Transmission/Receiver mode indicate active communications between all beacons. The first stage is the EBS approaching the Receiver drone, which is perpendicular to each other for the second stage. The third stage highlights the displacement due to the effects of downwash. These experiments help validate the accurate position to execute Aerobridge.

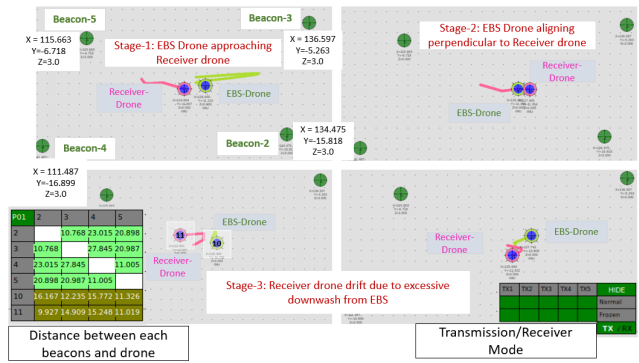

Figure 12: Indoor GPS Coordinate system to detect drones' coordinates and distance.

The [x,y] coordinates and the height measurement in meters are the three parameters output in the motion tracking system. The effect of downwash is more substantial immediately beneath the UAV and diminishes as one moves away from the center of the UAV. Fig. 13 demonstrates the results of the drone displacement due to the effect of the downwash. The video for this experiment is attached here (**Link**).

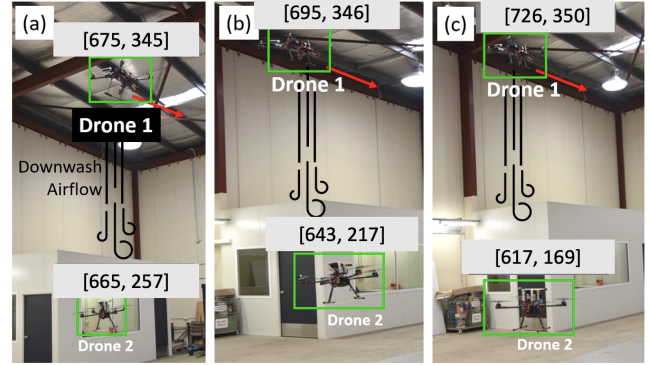

Figure 13: Drone displacement due to excessive airflow (a) 1st iteration (b) 2nd iteration (c) 3rd iteration.

It can be observed in Fig. 13 (a) that when the drone is positioned at a 3 m distance from each other, the receiver drone is exposed to the downwash from the EBS drone. Due to the downwash turbulence, the drone below is destabilized and can drift across the x or y axes. The horizontal displacement is captured in Fig. 13 (b) and (c). The altitude measurements were extracted from the flight. These measurements validate the optimal position and trajectories of the drone. Fig. 14 a shows the average displacement of the receiver drone from the EBS drone for different height separations. Here, the y-axis represents the displacement of the drones in meters, and the x-axis represents the distance between drones in meters. In the hover state, at an altitude ranging between 1 to 4 meters, the drone has a very high displacement that gradually decreases as the distance increases. The 50 iterations of drone flights demonstrate displacement of the drone positioned below (i.e. the Receiver drone) while being approached using four possible trajectories. The effect of that has been analyzed in Fig. 14 (b), where points 'H-V', 'H-H', 'V-V' and 'V-H' represent different trajectories 'Horizontal-then-Vertical', 'Horizontal-Horizontal', 'Vertical-Vertical', and 'Vertical-then-Horizontal' respectively for the EBS drone to approach the Receiver drone. The scatter plot highlights the displacement values in meters for each trajectory tested for 50 iterations. Trajectory 'H-H' demonstrates wherein the EBS drone first attempts horizontal alignment of 0.5 m between the two drones and then vertically aligns to 0.5 m 'alt', and Trajectory 'V-H' demonstrates the proposed trajectory approach wherein the EBS drone first attempts vertical alignment of 0.5m 'alt' and then aligns horizontally to 0.5 m distance. However, in 'H-H', both drones are moving to align themselves horizontally, assuming the vertical distance of 0.5 m has been achieved. Similarly, in 'V-V', the vertical alignment movement has been tested, assuming horizontally, the drones are aligned at a 0.5m distance. The trajectories for 'H-V' and 'H-H' have demonstrated the highest displacement between the drones that are caused due to the downwash, as it leads to several occurrences of more than 50% overlap due to the horizontal approach method, which results in airflow disturbance and change in thrust as demonstrated in Fig 6. However, trajectories 'V-V' and 'V-H' demonstrate minimal displacement, as the resultant overlap was always 50% or less. In the real-world testing it is not always possible to predetermine the horizontal alignment, these experiments indicate the 'Vertical-then-Horizontal' trajectory to be the most optimal for testing and verification; this allows the EBS drone to be closer Thus, it is not ideal to transfer between two drones while they are perpendicular to each other due to instability. The low vertical distance of 0.5 m between two drones allows for a balanced and quick transfer.

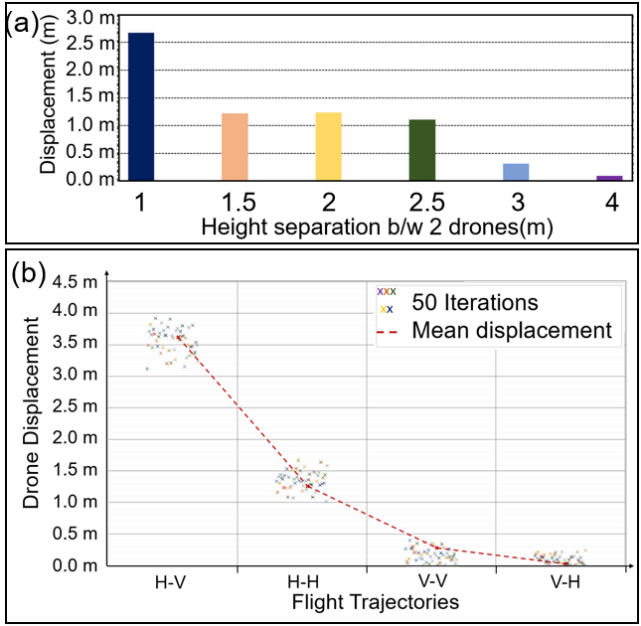

Figure 14: (a) Drone Disp at different height separations (b) Drone Displacement for different trajectories at 1.5 m (50 iterations).

## 6.2 CMP Drone Localization

In this section, we present the results of our experiments aimed at validating the accuracy of the proposed pose estimation system for the EBS drones using CMP localization. The system's ability to achieve accurate position locking while maintaining the desired orientation relative to the front marker is thoroughly evaluated. We conducted indoor experiments to validate the EBS drone positioning of the CMP containing Receiver drone set in different directions. The CMP detection method is illustrated in Fig. 15, detecting five markers with their ID information and the reference position and distance from the camera. The camera output is a default of 30 fps, and the time for detection is 1 ms.

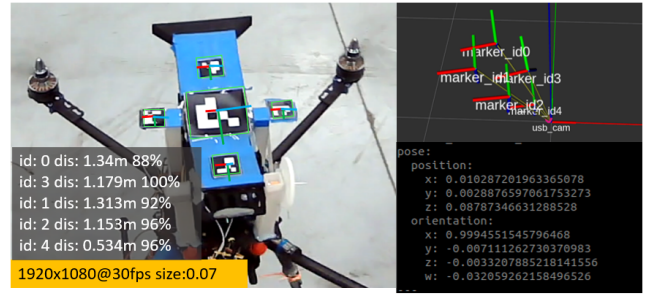

Figure 15: Unique marker position estimate for receiver drones with ROS 'tf' reference for each marker.

The CMP method can detect the 70 mm marker up to 3m in range. The depth camera positioned on the EBS drone modifies the camera's position and changes the marker's pose. The camera was tested for positions to determine the Field of View (FoV) detection range and the optimum tilt angle. The camera can detect the markers within a tilt range of 30 degrees to 55 degrees. The optimum angle for detection during the battery transfer is 45 degrees.

The ground truth positions were obtained through a motion tracking system to serve as a reference for our evaluations. We calculated the position error for each marker association to assess the accuracy of position locking. The position error was computed as the Euclidean distance between the drone's estimated position and the ground truth position of the designated marker. The equation derives the position error 'e' between the drone's estimated and actual ground truth positions. The 'e' values are approximately 0.9 cm on an average of 20 iterations, which indicates higher accuracy, implying the drone is effectively locked onto the intended marker position. The results demonstrate that the proposed system achieves highly accurate position lock, with errors consistently below a predefined threshold; this indicates that the drone successfully reaches the intended marker positions with minimal deviation. For each marker association $e.g., Left-Back, Right-Back, Front-Back$, the position error is calculated as follows, where x_est, y_est and z_est are estimated position across all axes and x_IP, y_IP and z_IP are the desired position values :

$$e = \sqrt{(x\_est - x\_IP)^2 + (y\_est - y\_IP)^2 + (z\_est - z\_IP)^2}$$
(9)

The orientation control aspect for the EBS drone was evaluated by analyzing the orientation deviation between the drone's estimated orientation and the desired orientation relative to the front position of the CMP design. This deviation was calculated as the angular difference between the two orientations. Fig. 16 presents the orientation deviation analysis results. The orientation deviation measures the angular difference between the drone's estimated and desired orientation relative to the front marker. It signifies how closely the drone's heading aligns with the specified orientation. For each marker association, the orientation deviation is calculated as follows, wherein $\theta_{e}st$ is the estimated drone angle and $\theta_I P$ is the desired position:

$$\alpha = |\theta\_est - \theta\_IP| \qquad (10)$$

This calculation yields the deviation value in $\alpha$, representing how much the drone's orientation deviates from the desired direction. The $\alpha$ measures < 1 degree over an average of 20 tests indicating that the drone successfully maintains the required position. Fig. 16 shows that the system effectively maintains the drone's orientation within a narrow range of deviation from the desired orientation; this validates the system's ability to correct its orientation accurately with approximately a degree of error and adhere to the specified orientation constraint. The red plot in Fig. 16 highlights orientation correction from the right to the back, the angle measurements vary between 30 to 50 degrees, and the green plot demonstrates angular correction between -30 to -50 degrees from the left position.

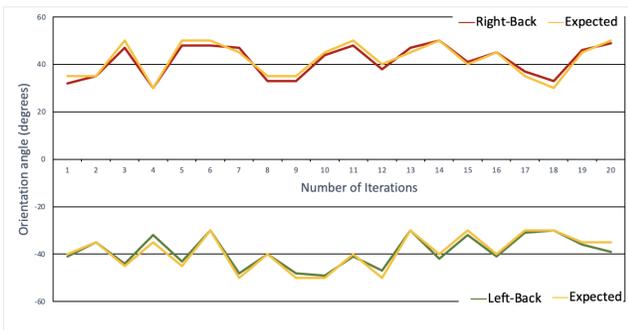

Figure 16: Orientation angle correction for Right and Left position for 20 iterations.

Table 1 summarizes each marker association's average position error and orientation deviation to provide a comprehensive system performance overview.

The equation derives the position error (e) between the drone's estimated and actual ground truth positions. The 'e' values are approximately 0.9 cm on an average of 20 iterations, which indicates higher accuracy, implying the drone is effectively locked onto the intended marker position. The results demonstrate that the proposed system achieves highly accurate position locking, with position errors consistently below a predefined threshold; this indicates that the drone successfully reaches the intended marker positions with minimal deviation. The results in Table 1 reinforce the consistent accuracy of the system across different marker associations. The low average position errors and orientation deviations further establish the system's capability to achieve accurate position locking while maintaining the correct orientation, validating the CMP design's effectiveness.

Table 1: CMP localization performance

| Marker Association | Avg Pos. error | Avg Ort dev. |
|---|---|---|
| Left-Back | 0.693 | 1.35 |
| Right-Back | 0.907 | 1.55 |
| Front-Back | 1.107 | 1.7 |

### 6.3 Battery Transfer Validation

Experiments were conducted to validate the proposed battery transfer system's overall effectiveness and feasibility. The experiments aimed to assess the performance of the visual-inertial navigation method in achieving precise alignment between the drones during mid-air battery transfers. A total of 15 outdoor flight tests or iterations were conducted to replicate different positions and flight dynamics. Ambient wind conditions were less than 4 m/s. Fig. 17 illustrates the position error for various marker associations. The purple plot line demonstrates the position correction from front to back; the drone varies the position between 4.5 to 15 cm. These measurements account for the distance between the markers in the CMP design to aid the drone's movement. The position error is approximately 0.9 cm on average, and it measures how accurately the drone's estimated position aligns with the ground truth position of the designated marker. The drone has demonstrated a successful transfer despite the offset of 0.9 cm.

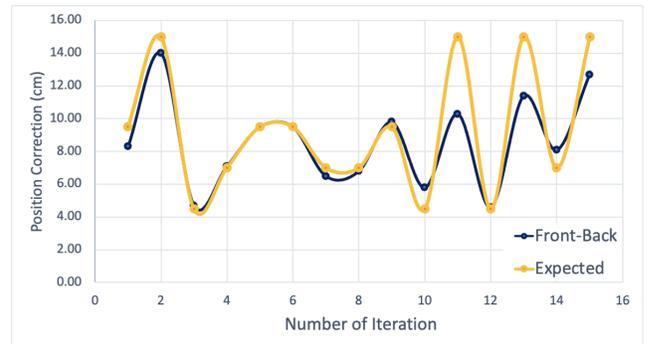

Figure 17: Position correction for the Front position for 15 iterations.

The battery transfer mechanism implemented in the EBS drone system enables a stable and efficient transfer process with 4m/s wind speed during the flight tests as seen in Fig. 18. The two drones can securely align with a 45-degree slide on the EBS drone measuring 230 mm and a corresponding slide on the receiver drone measuring 210 mm. The AeroBridge alignment arrangement ensures that the drones remain within an approximate distance of 0.5m during the transfer. The battery transfer is completed swiftly, taking on average of 5 seconds. The 5 seconds window consists of opening of the aerobridge, battery handoff and bridge close. In this example, we see in stage 1 of the figure the EBS drone with battery enclosed, stage 2 with both EBS and receiver drones with their slides extended and the battery being handed off, followed by stage 3 with the battery successfully transferred within the now folded up slide of the receiver drone.To estimate the smoothness of the transfer, we measured the amount of vibration encountered during the handoff. Fig. 19 shows the typical vibration along the x, y, and z axes experienced during one of the EBS and Receiver drone tests, in blue, orange, and green plots, respectively. The vibration has been measured for the transfer time between 3:30 to 3:40. The graphs indicate a stable handover as the vibrations across all axes are close to zero. The experimental data supported the notion that the angular approach, combined with visual-inertial odometry, effectively mitigated the challenges associated with direct downwash and ensured a stable and efficient battery transfer process. Our AeroBridge system incorporates a 45-degree transfer mechanism that leverages gravity for efficient battery transfer, sidestepping downdraft instability by diagonally placing the drones. This transfer mechanism is executed at 4m/s windspeed outdoors, and the transfer process was 4 seconds. These validation experiments provide strong evidence for the viability of the proposed system and pave the way for its practical implementation in real-world scenarios. Here is a video link of the outdoor implementation of AeroBridge (**Link**).

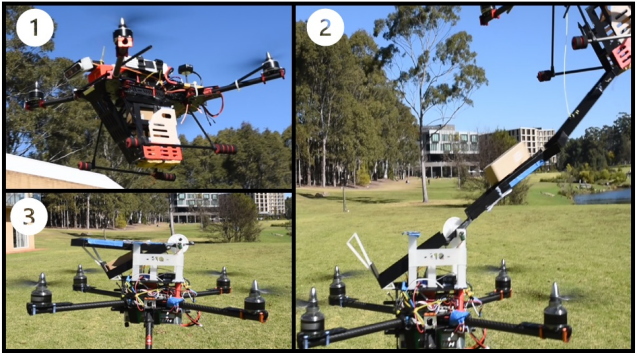

**Figure 18: Battery Handoff Process**

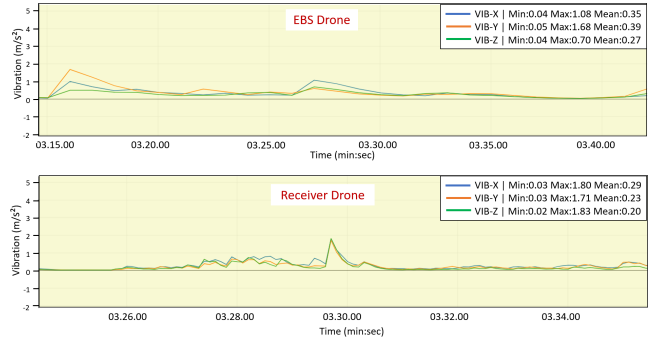

**Figure 19: Vibration across all axes for EBS and Receiver drone during transfer.**

Our outdoor experiments exposed the system to non-ideal wind conditions of up to 4 m/s, as depicted in Fig. 18. Despite such challenging environmental factors, our vision system navigated through sun glare conditions, ensuring little to no interruptions. Notably, the system exhibited a minimal latency of 1 ms in detection time, as detailed in Section 6.2, underscoring its responsiveness. Furthermore, our tests revealed an average localization accuracy of 0.9 cm, outlined in Section 6.3. These findings collectively show the potential and robustness of our developed system across dynamic conditions.

## 7 CONCLUSION

In conclusion, this paper presents an innovative solution to address the limitations of drone flight duration by proposing an EBS drone capable of transferring fresh batteries to depleted drones in mid-air. The AeroBridge system's diagonal slide transfer mechanism, capitalizing on gravity for efficient battery exchange, showcases a remarkable ability to overcome instability and achieve precise mid-air alignment. The experimental results validate the efficiency and stability of the battery transfer system, with successful transfers completed within 5 seconds and the drones maintaining a vertical distance of 0.5m during the process.

In future research, one primary focus would be the robustness of the EBS drone against harsh environmental conditions, such as strong winds. Future research will also center on evaluating system performance with an expanding number of drones needing battery swaps, identifying limitations, and optimizing scalability opportunities. Additionally, we will explore the feasibility of a two-way drone battery swap. There is also potential for extending AeroBridge to other multirotor drone platforms.